\documentclass[10pt,twocolumn,letterpaper]{article}

\usepackage[pagenumbers]{wacv} 

\usepackage{graphicx}
\usepackage{amsmath}
\usepackage{amssymb}
\usepackage{booktabs}

\usepackage{color,xcolor}
\usepackage{epsfig}
\usepackage{graphicx}

\usepackage{adjustbox}
\usepackage{array}
\usepackage{booktabs}
\usepackage{colortbl}
\usepackage{float,wrapfig}
\usepackage{hhline}
\usepackage{multirow}
\usepackage{subcaption} 
\usepackage[toc,page]{appendix}

\usepackage{amsmath,amsfonts,amsthm,amssymb}
\usepackage{bm}
\usepackage{nicefrac}
\usepackage{microtype}
\usepackage{inconsolata}
\usepackage{pifont}

\usepackage{changepage}
\usepackage{extramarks}
\usepackage{fancyhdr}
\usepackage{lastpage}
\usepackage{setspace}
\usepackage{soul}
\usepackage{xspace}
\usepackage{indentfirst}
\usepackage{paralist}
\usepackage{booktabs,arydshln}

\usepackage[pagebackref=true,breaklinks=true,colorlinks,citecolor=citecolor,bookmarks=false]{hyperref}

\usepackage{algorithm, algorithmic}
\usepackage{enumerate}
\usepackage{lipsum}

\DeclareMathOperator*{\argminA}{\,min}
\DeclareMathOperator*{\argmaxA}{\,max}

\newcolumntype{L}[1]{>{\raggedright\let\newline\\\arraybackslash\hspace{0pt}}m{#1}}
\newcolumntype{C}[1]{>{\centering\let\newline\\\arraybackslash\hspace{0pt}}m{#1}}
\newcolumntype{R}[1]{>{\raggedleft\let\newline\\\arraybackslash\hspace{0pt}}m{#1}}

\newcommand{\ignorethis}[1]{}

\makeatletter
\DeclareRobustCommand\onedot{\futurelet\@let@token\@onedot}
\def\@onedot{\ifx\@let@token.\else.\null\fi\xspace}

\makeatother

\makeatletter
\def\adl@drawiv#1#2#3{%
        \hskip.5\tabcolsep
        \xleaders#3{#2.5\@tempdimb #1{1}#2.5\@tempdimb}%
                #2\z@ plus1fil minus1fil\relax
        \hskip.5\tabcolsep}
\newcommand{\cdashlinelr}[1]{%
  \noalign{\vskip\aboverulesep
           \global\let\@dashdrawstore\adl@draw
           \global\let\adl@draw\adl@drawiv}
  \cdashline{#1}
  \noalign{\global\let\adl@draw\@dashdrawstore
           \vskip\belowrulesep}}
\makeatother

\definecolor{citecolor}{HTML}{0071bc}
\definecolor{mydarkblue}{rgb}{0,0.08,1}
\definecolor{mydarkgreen}{rgb}{0.02,0.6,0.02}
\definecolor{darkred}{rgb}{0.8,0.02,0.02}
\definecolor{darkorange}{rgb}{0.40,0.2,0.02}
\definecolor{darkpurple}{RGB}{111,0,255}
\definecolor{myred}{rgb}{1.0,0.0,0.0}
\definecolor{mygold}{rgb}{0.75,0.6,0.12}
\definecolor{mydarkgray}{rgb}{0.66, 0.66, 0.66}
\definecolor{myemerald}{RGB}{0, 152, 116}

\definecolor{spc}{RGB}{119, 107, 170}
\definecolor{pct}{rgb}{0.7, 0, 0.2}

\newcommand{\myparagraph}[1]{\vspace{-10pt}\paragraph{#1}}

\def\wacvPaperID{806} 
\def\confName{WACV}
\def\confYear{2024}

\usepackage[capitalize]{cleveref}
\crefname{section}{Sec.}{Secs.}
\Crefname{section}{Section}{Sections}
\Crefname{table}{Table}{Tables}
\crefname{table}{Tab.}{Tabs.}

\begin{document}

\title{CrAFT:~\ul{C}omp\ul{r}ession-\ul{A}ware \ul{F}ine-\ul{T}uning for Efficient Visual Task Adaptation}


\author{
Jung~Hwan~Heo\textsuperscript{1,2} \quad
Seyedarmin Azizi\textsuperscript{1} \quad
Arash Fayyazi\textsuperscript{1} \quad
Massoud Pedram\textsuperscript{1} \\
\textsuperscript{1}University of Southern California \hspace{10mm} \textsuperscript{2}NAVER Cloud \\
\small{\texttt{\{johnheo,seyedarm,fayyazi,mnazemi,pedram\}@usc.edu}}
}

\maketitle

\footnotetext{Correspondence to \texttt{johnheo@usc.edu}}

\begin{abstract}
\noindent Transfer learning has become a popular task adaptation method in the era of foundation models. However, many foundation models require large storage and computing resources, which makes off-the-shelf deployment impractical. Post-training compression techniques such as pruning and quantization can help lower deployment costs. Unfortunately, the resulting performance degradation limits the usability and benefits of such techniques. To close this performance gap, we propose \textbf{CrAFT}, a simple fine-tuning framework that enables \textbf{effective post-training network compression}. In CrAFT, users simply employ the default fine-tuning schedule along with \textbf{sharpness minimization} objective, simultaneously facilitating task adaptation and compression-friendliness. Contrary to the conventional sharpness minimization techniques, which are applied during pretraining, the CrAFT approach adds \textbf{negligible training overhead} as fine-tuning is done in under a couple of minutes or hours with a single GPU. The effectiveness of CrAFT, which is a general-purpose tool that can significantly boost one-shot pruning and post-training quantization, is demonstrated on both convolution-based and attention-based vision foundation models on a variety of target tasks. The code will be made publicly available.

\end{abstract}

\section{Introduction} 
\label{sec:intro}
\noindent Convolutional Neural Networks (CNNs) \cite{he2016deep, krizhevsky2017imagenet, tan2019efficientnet, liu2022convnet} and Vision Transformers (ViTs) \cite{dosovitskiy2020image, liu2021swin, he2022masked, dehghani2023scaling, touvron2021training} have brought massive empirical success in various vision tasks including image classification \cite{he2016deep}, object detection \cite{tan2020efficientdet}, semantic segmentation \cite{howard2019searching}, and more. With state-of-the-art pretraining techniques, self-supervised approaches such as MAE \cite{wortsman2022model} and billion parameter scale vision foundation models \cite{dehghani2023scaling} can achieve up to 90.94\% and 89.5\% Top-1 Accuracy on ImageNet-1k, respectively. However, the high computational cost challenges the development and deployment of neural networks in resource-constrained environments.

An undercurrent to the ever-scaling deep learning revolution is model compression research, where the goal is to squeeze large models into smaller compact ones without compromising their performance. Various methods have been developed, such as pruning \cite{han2015deep, he2017channel, frankle2018lottery, liu2018rethinking, yang2021nvit}, quantization \cite{nagel2019data, yang2019quantization, cho2021dkm, banner2019post}, and knowledge distillation\cite{hinton2015distilling, lopes2017data, wang2018dataset, beyer2022knowledge}. Thanks to the great progress, large language models such as GPT-3 \cite{brown2020language} and LLaMA \cite{touvron2023llama} can be pruned in one-shot (without retraining) to 50\% with negligible accuracy loss \cite{frantar2023massive, sun2023simple}.


One trend here is that the vast majority of model compression research focuses on \emph{post-training} methods, which does not account for the potential for better compression schemes \emph{during training itself}. This is mainly due to the astronomical pretraining costs that make it impractical for researchers to attempt such expensive projects. A few pretraining methods exist that optimize for compressibility, but they are limited to either pruning \cite{miao2021learning, peste2022cram} or quantization \cite{liu2021sharpness, liu2023binary}, thus are not suitable as a general compression framework. Given the simplicity and effectiveness of post-training compression, we aim to shift attention towards the contrary and ask: what is a \textbf{general and effective} \emph{training method} that can ultimately \textbf{improve the efficacy} of \emph{post-training compression}?

In this paper, we present a simple yet effective fine-tuning framework that is inspired by the sharpness-minimizing property. To achieve this, we formulate the problem of model compression into compressibility and generalizability and distill it down to a unified view of loss landscapes. We show the efficacy of our approach in a wide range of \textit{training-free} compression methods, including one-shot weight pruning, INT8/INT6 quantization, and weight-only quantization. 


\begin{itemize}

    \item We conduct a detailed analysis of the relationship between loss sharpness and compressibility. For the first time, we show that sharpness-minimized models have low intrinsic rank and smooth activation ranges. This motivates the sharpness-minimization objective during fine-tuning to enable better post-training compression.
    
    \item We present \textbf{CrAFT}, a compression-aware fine-tuning framework that enables \textbf{compressed and task adapted} vision models. Since our approach is orthogonal to existing compression techniques, we demonstrate a significant accuracy boost when adopting CrAFT.

    \item We extensively test our framework by applying to a wide range of network backbones, including ViTs and CNNs on various transfer learning tasks (e.g., Flowers, Cars, Food). CrAFT can lift the accuracy by up to \textcolor{myemerald!80}{\textbf{12\%}} and \textcolor{myemerald!80}{\textbf{3.3\%}} on pruning and quantization, respectively. 
\end{itemize}
\section{What Makes Compression Difficult?}\label{sec:motiv}

\textbf{Preliminaries.}
\noindent Recently, sharpness-aware minimization (SAM) techniques have been shown to optimize for generalizability but also compressibility in both language modeling and image classification tasks \cite{na2022train, heo2023fast}. However, the source of such compressibility is not well understood. In this section, we investigate the sources of compressibility challenges for pruning and quantization using an off-the-shelf ViT-B model from the Timm library \cite{rw2019timm} and see how SAM alleviates them. We unify the compression difficulties into the perspective of a loss geometry $\lambda_\mathrm{max}$, motivating our method for CrAFT. Here, $\lambda_\mathrm{max}$ represents the \textit{worst curvature of the loss }which is equivalent to the maximum eigenvalue of the Hessian matrix (convexity of loss surface).

\subsection{Pruning Difficulty}
\label{subsec:prune-difficult}
\begin{figure}[!h]
    \centering
    \includegraphics[width=\linewidth]{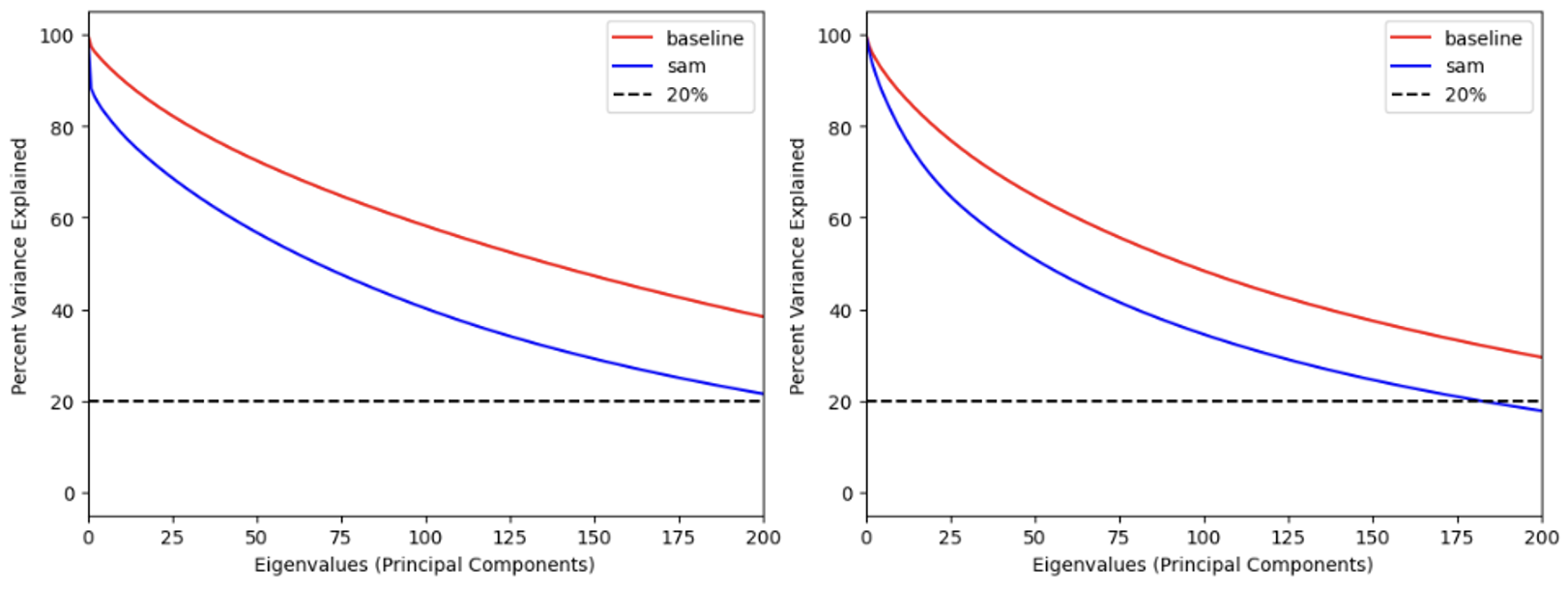}
    \caption{\textbf{Eigenvalue Decomposition.} ViT-B trained with the SAM optimizer results in weights with lower inherent dimensions. Averaged over all 12 blocks, both Feed-Forward Networks (Left) and QKV Attention Matrices (Right) show that 80\% or more of the weight variance (entropy) can be constructed by the first 200 principal components out of the 768 full channel size.}
    \label{fig:motiv-prune}
\end{figure}
\noindent Sparsity is a long-standing regularization strategy that seeks to find the simplest solution that can generalize well outside the training set. Hence, the pruning task can be cast as a constrained optimization problem to find the sparse subnetworks that are more parameter-efficient without degrading the performance. 

Given that most networks are overparameterized, LoRA \cite{hu2021lora} uses a small number of parameters (low-rank adaptors) to fine-tune the larger model. Similarly, we quantify parameter efficiency through Principal Component Analysis.

As shown in Figure~\ref{fig:motiv-prune}, the SAM optimizer produces weights that have an inherently lower rank. Specifically, 80\% of the information entropy of the Feedforward Network weights can be explained by 200 principal components whereas the same number of principal components account for only 40\% of the information when a baseline SGD optimizer is used. For QKV attention weights, the rank is even smaller. We can learn from this that explicitly minimizing the sharpness of the loss landscape fundamentally eases the pruning difficulty. For example, in the one-shot pruning scenario and at 50\% sparsity level, SAM retains 3.6\% more accuracy compared to a baseline model optimized with ADAM (Table~\ref{tab:motiv:sam}).

\begin{table}[!h]
    \setlength{\tabcolsep}{8pt}
    \small\centering
    \caption{\textbf{Compressibility of SAM.} ViT pretrained with SAM consistently demonstrates superior robustness to various model compression schemes compared to ADAM baseline. ADAM additionally uses heavy augmentation (AugReg \cite{steiner2021train}), which leads to higher performance for pretrained Top-1 accuracy on ImageNet-1k. Speedup is measured on NVIDIA A6000 GPU.}
    \begin{tabular}{lcc}
        \toprule
        ViT-B   & ADAM      & SAM \\
        \midrule
        $\lambda_\mathrm{max}$   & 739 & 21 \\
        Pretrained Accuracy & 84.58 & 80.28 \\
        \midrule
        Weight Pruning (50\% Sparsity) & -4.78 & \textbf{-1.11}\\
        Token Pruning (1.18$\times$ Speedup) & -4.30 & \textbf{-1.04}\\
        Token Merging (1.90$\times$ Speedup) & -4.11 & \textbf{-0.95}\\
        {A8W8 Uniform Quantization} & -72.05 & \textbf{-0.15} \\
        \bottomrule
    \end{tabular}
    \label{tab:motiv:sam}
    \vspace{-8pt}
\end{table}

\subsection{Quantization Difficulty}
\label{subsec:quant-difficult}
\begin{figure}[!h]
    \centering
    \includegraphics[width=\linewidth]{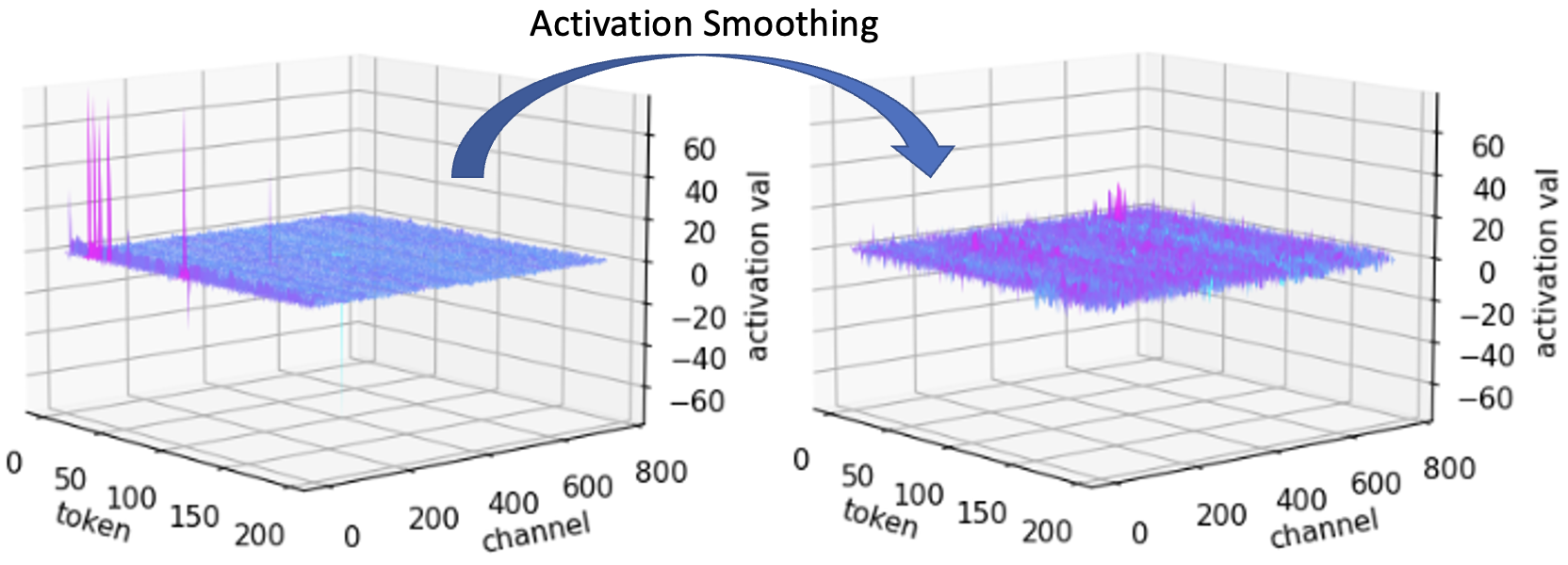}
    \caption{\textbf{Activation Smoothing.} Vanilla ViT-B (left) suffers from activation outliers (large magnitude) while ViT-B-SAM (right) enjoys a smooth distribution. With smaller dynamic range, the quantization difficulty is easier for SAM-trained models.}
    \label{fig:motiv-quant}
    \vspace{-8pt}
\end{figure}
\begin{figure*}[!t]
    \small\centering
    \includegraphics[width=0.75\paperwidth]{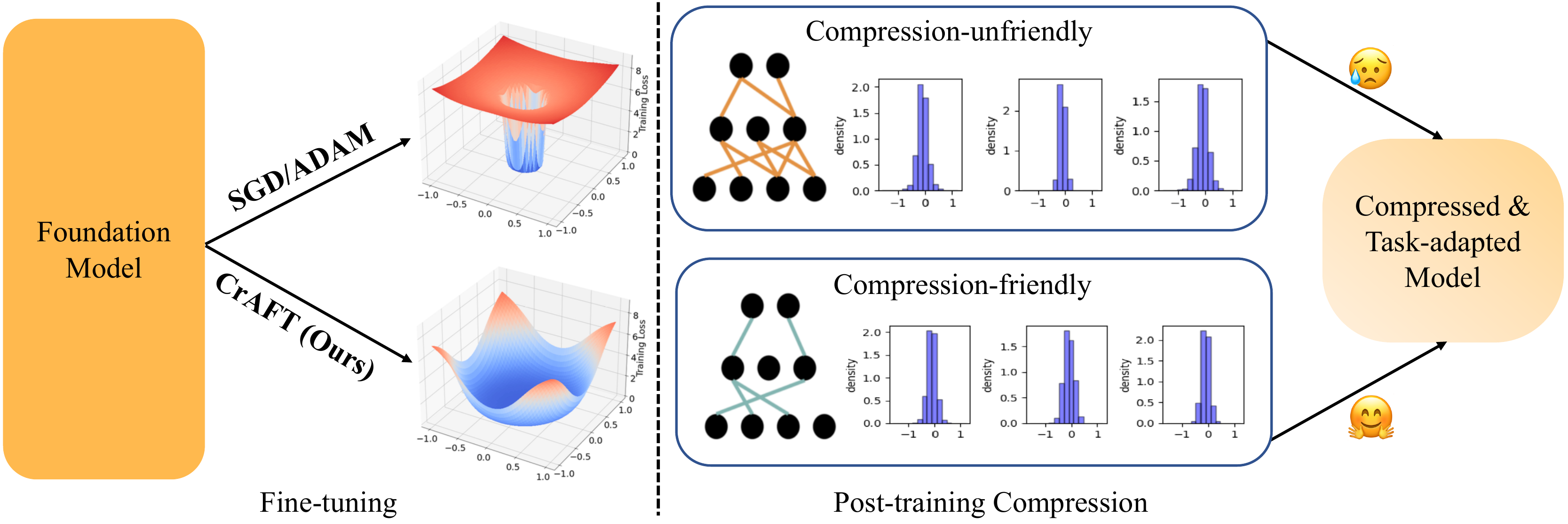}
    \caption{\textbf{CrAFT is a Compression-Aware Fine-Tuning framework} that enables \textit{well-compressed and task adapted} vision foundation models. By formulating the network \textit{compression problem} into \textit{generalization problem}, CrAFT applies  sharpness minimization during fine-tuning to obtain networks with flatter loss landscapes (small $\lambda_\mathrm{max}$).  With higher inherent sparsity and smoother activations, it delivers superior results to existing post-training compression methods such as one-shot pruning and PTQ. }
    \label{fig:teaser}
\vspace{-8pt}
\end{figure*}
\noindent Large language models (LLMs) have activation outliers in certain channels where the increased dynamic range makes it very difficult to apply static quantization \cite{dettmers2022llm, xiao2022smoothquant}. Similarly, in Figure~\ref{fig:motiv-quant}, we observe the "activation outlier problem" in vanilla ViT-B. The first channel spans from $-60$ to $60$ whereas ViT-B-SAM leads to a smoother activation distribution. In the A8W8 quantization scenario (Table 2), SAM can drastically improve the quantization robustness with a \textbf{mere 0.15\% loss} while ADAM degrades it by 72.05\%!

We acknowledge that quantization-specific solutions such as mixed-precision quantization \cite{dettmers2022llm} or equalization methods \cite{nagel2019data,xiao2022smoothquant} can be adopted in ViT as well. Yet, we aim for generality across different compression schemes, so we focus on minimizing the $\lambda_\mathrm{max}$ as the core compression difficulty metric. 

\subsection{Optimization Difficulty}
\label{subsec:optim-difficult}
\noindent As stated earlier, although SAM can effectively minimize the sharpness of loss surface, it significantly slows down training by up to 2$\times$ \cite{duefficient,dusharpness} due to the required second-order computations. This can be a blocker for adopting pretraining techniques for compression, as even small overheads can cause a considerable budget increase in large-scale pretraining scenarios. We identify this as an optimization difficulty and purposely avoid expensive optimization strategies. 

To summarize, an ideal method for network compression should (i) be generalizable across network backbones and compression schemes, (ii) not incur significant training overheads that hamper practical adoption, and (iii) be effective in driving down $\lambda_\mathrm{max}$ to encourage ease of post-training compression. Given these design criteria, we now present our proposed CrAFT framework.

\section{Fine-tuning with Sharpness Minimization}
\label{sec:method}

This section explains sharpness aware minimization (SAM) and how we use it in fine-tuning vision foundation models, resulting in a relatively flat loss landscape while adding little to no training overhead. Since this fine-tuning is orthogonal to post-training compression techniques, it can significantly boost accuracy. In general, our focus is to comprehend the following:
Is it true that models located in flatter minima can be compressed more easily?

SAM has been shown to not only optimize for generalizability but also compressibility in both language modeling and image classification tasks \cite{na2022train, heo2023fast}. However, the source of such compressibility is not well understood. In this section, we investigate the sources of compressibility challenges for pruning and quantization and see how SAM alleviates them. We unify the difficulties into the perspective of a loss geometry $\lambda_\mathrm{max}$, laying the foundation for our proposed CrAFT framework. 

\subsection{Sharpness Aware Minimization}

The Hessian of the loss function with respect to weights signifies the convexity and sharpness of the loss surface. Thus, we can take the worst curvature of the loss by taking the maximum eigenvalue of the Hessian matrix, which we denote as $\lambda_\mathrm{max}$ henceforth. 

Instead of finding a specific location with a local minimum loss, Sharpness-Aware Minimization (SAM) aims to find a region where the training loss is consistently low. This region should have both low loss values and low curvature. This objective can be expressed as a min-max optimization problem as follows:
\begin{equation}
\argminA_w \argmaxA_{||\epsilon||_2 \leq \rho}  \mathcal{L} (w + \epsilon)
\label{eq:sam}
\end{equation}
The goal of the inner optimization problem is to find perturbations $\epsilon$ within a Euclidean ball of radius $\rho$. The aim is to maximize the perturbed loss $\mathcal{L} (w + \epsilon)$ by finding the optimal perturbations within this ball using a first-order Taylor expansion as 
\begin{equation}
\begin{aligned}
\hat{\boldsymbol{\epsilon}} & =\arg \max _{\|\epsilon\|_{2} \leq \rho} \mathcal{L}(\mathbf{w}+\boldsymbol{\epsilon}) \\ \\
& \approx \arg \max _{\|\epsilon\|_{2} \leq \rho} \mathcal{L}(\mathbf{w})+\boldsymbol{\epsilon}^{\top} \nabla_{\mathbf{w}} \mathcal{L}(\mathbf{w}) \\ \\
& \approx \rho \frac{\nabla_{\mathbf{w}} \mathcal{L}(\mathbf{w})}{\left\|\nabla_{\mathbf{w}} \mathcal{L}(\mathbf{w})\right\|_{2}} .
\end{aligned}
\label{eq:taylor}
\end{equation}
By substituting equation \ref{eq:taylor} back into equation \ref{eq:sam}, we then have the following optimization problem:

\begin{equation}
    \min _{\mathbf{w}} \mathcal{L}(\mathbf{w}+\hat{\boldsymbol{\epsilon}})
\end{equation}
As a final step, SAM adjusts the model weights by considering the gradient:
\begin{equation}
    \nabla_{\mathbf{w}} \mathcal{L}(\mathbf{w})|_{\mathbf{w}+\hat{\epsilon}}
\end{equation}

\subsection{Building Compression-friendliness}
Reference \cite{hochreiter1997long} introduced the idea that a flatter local minimum may improve the generalization performance of neural networks. Many subsequent works have explored the connection between loss geometry and generalization; for example, the Sharpness-Aware Minimizer (SAM) \cite{foret2020sharpness} explicitly minimizes the sharpness of the loss surface during training. However, due to the additional second-order computations, the training cost (and time) can go up by 2$\times$ \cite{duefficient}. 

We choose the vanilla SAM optimizer as the flagship representative of the sharpness-minimizing methods in fine-tuning phase and then evaluate the relationship between the loss of surface curvature and compressibility. We primarily focus on exploring compressibility within the context of post-training quantization \cite{PTQ4ViT_arixv2022}. In addition, we explore one-shot pruning approaches such as a simple magnitude-based pruning \cite{han2015deep} and activation-aware pruning \cite{sun2023simple}.

\section{Experiments}
\label{sect:exp}

\begin{table*}[!t]
    \small\centering
    \caption{\textbf{PTQ after Task Adaptation.} Results of ViT-B@IMNET-21k fine-tuned to various transfer learning datasets. As a simple baseline, we evaluate BasePTQ which uses symmetric uniform quantization \cite{DBLP:journals/corr/ZhouNZWWZ16}. More sophisticated PTQ4ViT \cite{PTQ4ViT_arixv2022} employs twin uniform quantization and hessian-guided scaling. We report absolute top1 accuracy (\%) and and the corresponding accuracy drop from full precision in parenthesis to the right. WxAy represents quantization configuration where x and y represent the number of bits for weights and activations, respectively. CrAFT consistently and often significantly boosts PTQ results compared to standard ADAM fine-tuning.}
    
    \begin{tabular}{lccccccc}
    \toprule
    Downstream Task & Method & FP32 & W8A8 & W6A6 & W6A16 & W5A16 & W4A16 \\
    \midrule
    \multirow{4}{*}{Flowers-102} & BasePTQ & 98.29 & 97.90 (-0.39) & 90.11 (-8.18) & 97.58 (-0.71) & 94.37 (-3.92)& 23.5 (-74.79) \\
    &  \cellcolor{gray!15}$+$ CrAFT & \cellcolor{gray!15}99.13 & \cellcolor{gray!15}98.94 (-0.19) & \cellcolor{gray!15}97.02 (-2.11)& \cellcolor{gray!15}\textbf{98.80 (-0.33)} & \cellcolor{gray!15}\textbf{98.03 (-1.10)} & \cellcolor{gray!15}69.48 (-29.66) \\
    & PTQ4ViT & 98.29 & 98.02 (-0.27) & 92.99 (-5.30) & 95.85 (-2.44) & 89.04 (-9.25) & 66.69 (-31.60) \\
    &\cellcolor{gray!15}  $+$ CrAFT &\cellcolor{gray!15} 99.13 &\cellcolor{gray!15} \textbf{99.11 (-0.02)} &\cellcolor{gray!15} \textbf{97.35 (-1.78)} &\cellcolor{gray!15} 98.39 (-0.74) &\cellcolor{gray!15} 96.24 (-2.89) &\cellcolor{gray!15} \textbf{85.71 (-13.43)} \\
    
    \midrule
    \multirow{4}{*}{Stanford Cars} & BasePTQ & 84.37 & 82.95 (-1.42) & 63.45 (-20.92) & 81.53 (-2.84) & 70.39 (-13.99) & 1.72 (-82.65) \\
    &  \cellcolor{gray!15}$+$ CrAFT & \cellcolor{gray!15}84.77 & \cellcolor{gray!15}84.12 (-0.65) & \cellcolor{gray!15}71.42 (-13.35)& \cellcolor{gray!15}\textbf{82.86 (-1.91)} & \cellcolor{gray!15}\textbf{77.11 (-7.66)} & \cellcolor{gray!15}3.37 (-81.40) \\
    & PTQ4ViT & 84.37 & 82.95 (-1.42) & 63.45 (-20.92) & 81.53 (-2.84) & 70.39 (-13.98) & 1.72 (-82.65) \\
    &\cellcolor{gray!15}  $+$ CrAFT &\cellcolor{gray!15} 84.77 &\cellcolor{gray!15} \textbf{84.47 (-0.30)} &\cellcolor{gray!15} \textbf{79.70 (-5.07)} &\cellcolor{gray!15} 82.32 (-2.44) &\cellcolor{gray!15} 76.93 (-7.84) &\cellcolor{gray!15} \textbf{49.55 (-35.22)} \\

    \midrule
    \multirow{4}{*}{Oxford Pets} & BasePTQ & 92.50 & 90.41 (-2.09) & 73.56 (-18.94) & 90.71 (-1.79) & 86.29 (-6.21)& 10.44 (-82.06) \\
    &  \cellcolor{gray!15}$+$ CrAFT & \cellcolor{gray!15} 93.02 & \cellcolor{gray!15}91.52 (-1.49) & \cellcolor{gray!15}84.47 (-8.56)& \cellcolor{gray!15}91.82 (-1.20) & \cellcolor{gray!15}\textbf{89.45 (-3.57)} & \cellcolor{gray!15}37.31 (-55.71) \\
    & PTQ4ViT & 92.50 & 91.82 (-0.68) & 87.98 (-4.52) &\cellcolor{myemerald!30} 88.31 (-4.20) & 79.75 (-12.75) & 46.33 (-46.17) \\
    &\cellcolor{gray!15}  $+$ CrAFT &\cellcolor{gray!15} 93.02 &\cellcolor{gray!15} \textbf{92.78 (-0.24)} &\cellcolor{gray!15} \textbf{89.99 (-3.02)} &\cellcolor{myemerald!30} \textbf{92.09 (-0.92)} &\cellcolor{gray!15} 89.23 (-3.78) &\cellcolor{gray!15} \textbf{74.27 (-18.75)} \\

    \midrule
    \multirow{4}{*}{Aircraft} & BasePTQ & 81.21 & 80.35 (-0.86) & 69.69 (-11.51) & 79.39 (-1.82) & 73.87 (-7.34)& 19.23 (-61.98) \\
    &  \cellcolor{gray!15}$+$ CrAFT & \cellcolor{gray!15} 81.24 & \cellcolor{gray!15}80.29 (-0.95) & \cellcolor{gray!15}74.38 (-6.86)& \cellcolor{gray!15}\textbf{79.78 (-1.46)} & \cellcolor{gray!15}75.99 (-5.24) & \cellcolor{gray!15}36.06 (-45.18) \\
    & PTQ4ViT & 81.21 & 80.56 (-0.65) & 75.31 (-5.90) & 78.76 (-2.45) & 75.24 (-5.96) & 57.61 (-23.60) \\
    &\cellcolor{gray!15}  $+$ CrAFT &\cellcolor{gray!15} 81.24 &\cellcolor{gray!15} \textbf{81.43 (+0.19)} &\cellcolor{gray!15} \textbf{78.33 (-2.90)} &\cellcolor{gray!15} 79.54 (-1.70) &\cellcolor{gray!15} \textbf{77.47 (-3.77)} &\cellcolor{gray!15} \textbf{68.41 (-12.83)} \\

    \midrule
    \multirow{4}{*}{Food} & BasePTQ & 89.78 & 89.11 (-0.67) & 73.92 (-15.86) & 87.69 (-2.08) & 78.71 (-11.07)& 2.82 (-86.95) \\
    &  \cellcolor{gray!15}$+$ CrAFT & \cellcolor{gray!15}90.80 & \cellcolor{gray!15} 90.26 (-0.54) & \cellcolor{gray!15} 81.67 (-9.13)& \cellcolor{gray!15} 89.24 (-1.56) & \cellcolor{gray!15} 84.53 (-6.27) & \cellcolor{gray!15} 11.09 (-79.70) \\
    & PTQ4ViT & 89.78 & 88.89 (-0.89) & 83.43 (-6.35) & 87.91 (-1.87) & 80.72 (-9.06) & 57.90 (-31.88) \\
    &\cellcolor{gray!15}  $+$ CrAFT &\cellcolor{gray!15} 90.80 &\cellcolor{gray!15} \textbf{90.45 (-0.35)} &\cellcolor{gray!15} \textbf{87.35 (-3.45)} &\cellcolor{gray!15} \textbf{89.48 (-1.32)} &\cellcolor{gray!15} \textbf{86.08 (-4.71)} &\cellcolor{gray!15} \textbf{75.57 (-15.23)} \\

    \midrule
    \multirow{4}{*}{CIFAR-100} & BasePTQ & 90.78 & 89.03 (-1.75) & 37.74 (-53.04) & 86.84 (-3.94) & 71.11 (-19.67)& 1.72 (-89.06) \\
    &  \cellcolor{gray!15}$+$ CrAFT & \cellcolor{gray!15} 89.74 & \cellcolor{gray!15} 88.89 (-0.85) & \cellcolor{gray!15} 78.77 (-10.97)& \cellcolor{gray!15} \textbf{88.13 (-1.61)} & \cellcolor{gray!15} 81.70 (-8.04) & \cellcolor{gray!15} 10.06 (-79.68) \\
    & PTQ4ViT & 90.78 & 89.79 (-0.99) & 79.14 (-11.64) & 87.04 (-3.74) & 74.08 (-16.70) & 32.21 (-58.57) \\
    &\cellcolor{gray!15}  $+$ CrAFT &\cellcolor{gray!15} 89.74 &\cellcolor{gray!15} \textbf{89.23 (-0.51)} &\cellcolor{gray!15} \textbf{85.40 (-4.34)} &\cellcolor{gray!15} 87.47 (-2.27) &\cellcolor{gray!15} \textbf{83.99 (-5.75)} &\cellcolor{gray!15} \textbf{68.76 (-20.98)} \\

    \midrule
    \multirow{4}{*}{CIFAR-10} & BasePTQ & 97.36 & \textbf{97.14} (-0.22) & 87.78 (-9.58) & 94.17 (-3.19) & 38.20 (-59.16) & 10.44 (-86.92) \\
    &  \cellcolor{gray!15}$+$ CrAFT & \cellcolor{gray!15} 97.30 & \cellcolor{gray!15}97.12 \textbf{(-0.18)} & \cellcolor{gray!15} 91.50 (-5.80)& \cellcolor{gray!15} 95.51 (-1.79) & \cellcolor{gray!15} 55.88 (-41.42) & \cellcolor{gray!15} 10.02 (-87.28) \\
    & PTQ4ViT & 97.36 & 97.11 (-0.25) & 89.72 (-7.64) & 95.27 (-2.09) & 67.10 (-30.62) & 12.5 (-84.86) \\
    &\cellcolor{gray!15}  $+$ CrAFT &\cellcolor{gray!15} 97.30 
    &\cellcolor{gray!15} 97.05 (-0.25) &\cellcolor{gray!15} \textbf{94.69 (-2.61)} &\cellcolor{gray!15} \textbf{96.23 (-1.07) }&\cellcolor{gray!15} \textbf{79.73 (-17.57)} &\cellcolor{gray!15} \textbf{23.5 (-73.80)} \\
    
    \bottomrule
    \end{tabular}
    \vspace{-8pt}
    \label{tab:results:transfer}
\end{table*}

\begin{table}[!t]
    \setlength{\tabcolsep}{3pt}
    \small\centering
    \caption{\textbf{Various Foundation Models.} Results of PTQ on both attention-based and convolution-based networks. @ denotes its pretrained dataset. All models are fine-tuned on CIFAR-100. Evaluations are done on the BasePTQ \cite{DBLP:journals/corr/ZhouNZWWZ16} method. CrAFT's compression-friendliness generalizes well across various architectures.}
        
    \begin{tabular}{lcccc}
        \toprule
        Backbone & Method & FP32 & W8A8 & W6A16 \\
        
        \midrule
        \multirow{2}{*}{ViT-B@IMNET-21k} & ADAM & \textbf{90.78} & 54.91 & 79.05 \\
        & \cellcolor{gray!15}$+$ CrAFT & \cellcolor{gray!15}89.74 & \cellcolor{gray!15}\textbf{81.84} & \cellcolor{gray!15}\textbf{85.76} \\
    
        \midrule
        \multirow{2}{*}{EfficientNet-B0@IMNET-1k} & ADAM & 84.75 & 77.32 & 78.39 \\
        & \cellcolor{gray!15}$+$ CrAFT & \cellcolor{gray!15}\textbf{85.51} & \cellcolor{gray!15}\textbf{81.43} & \cellcolor{gray!15}\textbf{82.53} \\
    
        \midrule
        \multirow{2}{*}{ResNet101@IMNET-1k} & ADAM & 83.82 & 73.32 & 82.66 \\
        & \cellcolor{gray!15}$+$ CrAFT & \cellcolor{gray!15}\textbf{84.18} & \cellcolor{gray!15}\textbf{81.12} & \cellcolor{gray!15}\textbf{82.96} \\
    
        \midrule
        \multirow{2}{*}{MobileNet-V2@IMNET-1k} & ADAM & 77.43 & 65.34 & 61.10 \\
        & \cellcolor{gray!15}$+$ CrAFT & \cellcolor{gray!15}\textbf{80.15} & \cellcolor{gray!15}\textbf{74.50} & \cellcolor{gray!15}\textbf{71.24} \\
        \bottomrule
    \end{tabular}
    \label{tab:results:backbone}
    \vspace{-8pt}
\end{table}

\subsection{Setup}

\paragraph{Dataset.}

Following commonly used evaluation setting for transfer learning \cite{kornblith2019better}, we use ImageNet \cite{russakovsky2015imagenet} as the pre-training dataset and then transfer the publicly available pre-trained model checkpoints from Timm \cite{rw2019timm} to 7 downstream object classification tasks, including Flowers \cite{DBLP:conf/icvgip/NilsbackZ08}, Cars \cite{DBLP:conf/iccvw/Krause0DF13}, Pets \cite{DBLP:conf/cvpr/ParkhiVZJ12}, Aircraft \cite{DBLP:journals/corr/MajiRKBV13}, Food \cite{DBLP:conf/eccv/BossardGG14}, CIFAR10 \cite{CIFAR10}, and CIFAR100 \cite{CIFAR10}. We conduct most of our evaluations with ViT-B@IMNET-21k, which uses ImageNet-21k as the pre-training dataset. All other models mentioned in this paper are pre-trained in the ImageNet-1k dataset.

\myparagraph{Models.}

To evaluate whether CrAFT generalizes across different network architectures, we use both attention-based and convolution-based models, including ViT-B, EfficientNet-B0, ResNet101, and MobileNet-V2. Again, most of the experiments are conducted with ViT-B, given the increasing popularity and efficacy of attention-based models in large-scale pre-training \cite{dosovitskiy2020image, he2022masked}. Since the scalability of base model architecture is necessary for forming a strong foundation model, we validate our method most heavily by targeting transformer-based foundation models. 

\myparagraph{Training Details.}

We fine-tune the entire network without conventional approaches, such as freezing the feature extractor. Because compression will be applied to the whole network, we seek to update all the parameters with the sharpness minimization objective to build compressibility. We leave more parameter-efficient fine-tuning (PEFT) schemes to future work, as the goal of the paper is to establish a strong baseline. Moreover, vision foundation models have relatively low memory consumption compared to their language counterparts. For example, the largest model in our paper, ViT-B, has 86M parameters, while today's popular language foundation model LLaMa \cite{touvron2023llama} has 7 billion as their smallest model.

The models are fine-tuned for five epochs using the Adam optimizer as a baseline and the SAM optimizer as part of the CrAFT framework. 
With batch size eight on a single NVIDIA A6000 GPU. The initial learning rate is tuned for each dataset while cosine schedule \cite{DBLP:conf/iclr/LoshchilovH17} is adopted for learning rate decay. 
For simplicity, we assume the batch size is 8 for all compared methods and datasets throughout the experiment section except for CIFAR experiments in which batch size is set to 64.

\subsection{Main Results}
For all our experiments, we compare CrAFT to ADAM fine-tuned models. After fine-tuning, we compare the quantization and pruning performance of the two fine-tuning methods. 

\subsubsection{Post-Training Quantization}

\paragraph{Baselines.}

As a strong baseline, we evaluate BasePTQ, which uses symmetric uniform quantization \cite{DBLP:journals/corr/ZhouNZWWZ16}. A more sophisticated scheme, PTQ4ViT \cite{PTQ4ViT_arixv2022}, employs twin uniform quantization by accounting for activation distributions in attention and feedforward blocks, as well as hessian-guided scaling to minimize the quantization error. 

\myparagraph{Transfer Learning Results.}

As in Table~\ref{tab:results:transfer}, CrAFT achieves consistent performance improvements over both BasePTQ and PTQ4ViT methods with much higher accuracy. For the Oxford Pets dataset, weight-only 6-bit quantization can lift the accuracy by 3\%, allowing $<$ 1\% degradation from full precision. 

We observe that PTQ4ViT is generally stronger in INT-based quantization, where both weights and activations are quantized. This is because its methodology includes activation-specific tricks that do not necessarily benefit the weight-only quantization scenario. This leads to BasePTQ being a strong baseline, outperforming in certain downstream tasks like Flowers-102 and Stanford Cars. However, PTQ4ViT shows superior performance in other domains, thanks to its hessian-based quantization scale search that accounts for additional second-order information. 

For more extreme weight-only quantization at 4 bits, CrAFT's effectiveness is magnified. Compared to its ADAM counterpart, it consistently improves performance by a large margin, from as low as 1\% (ref. Cars, BasePTQ) to as high as 46\% (ref. Flowers, BasePTQ). Yet, we recognize that these metrics can be fragile when the dataset is small, such as Flowers-102. Therefore, larger datasets such as Food have moderate accuracy gaps, whereas CrAFT has an accuracy boost of 18\%. 

Most surprisingly, CrAFT can actually bring an absolute accuracy improvement even when quantized. For W8A8 PTQ4ViT quantization on the Aircraft dataset, it has around 0.2\% improvement. We suspect this is potentially due to some regularization effect. 

\begin{figure*} [!tb]
\centering
    \begin{subfigure}{0.45\textwidth}
      \centering
      \includegraphics[width=\textwidth]{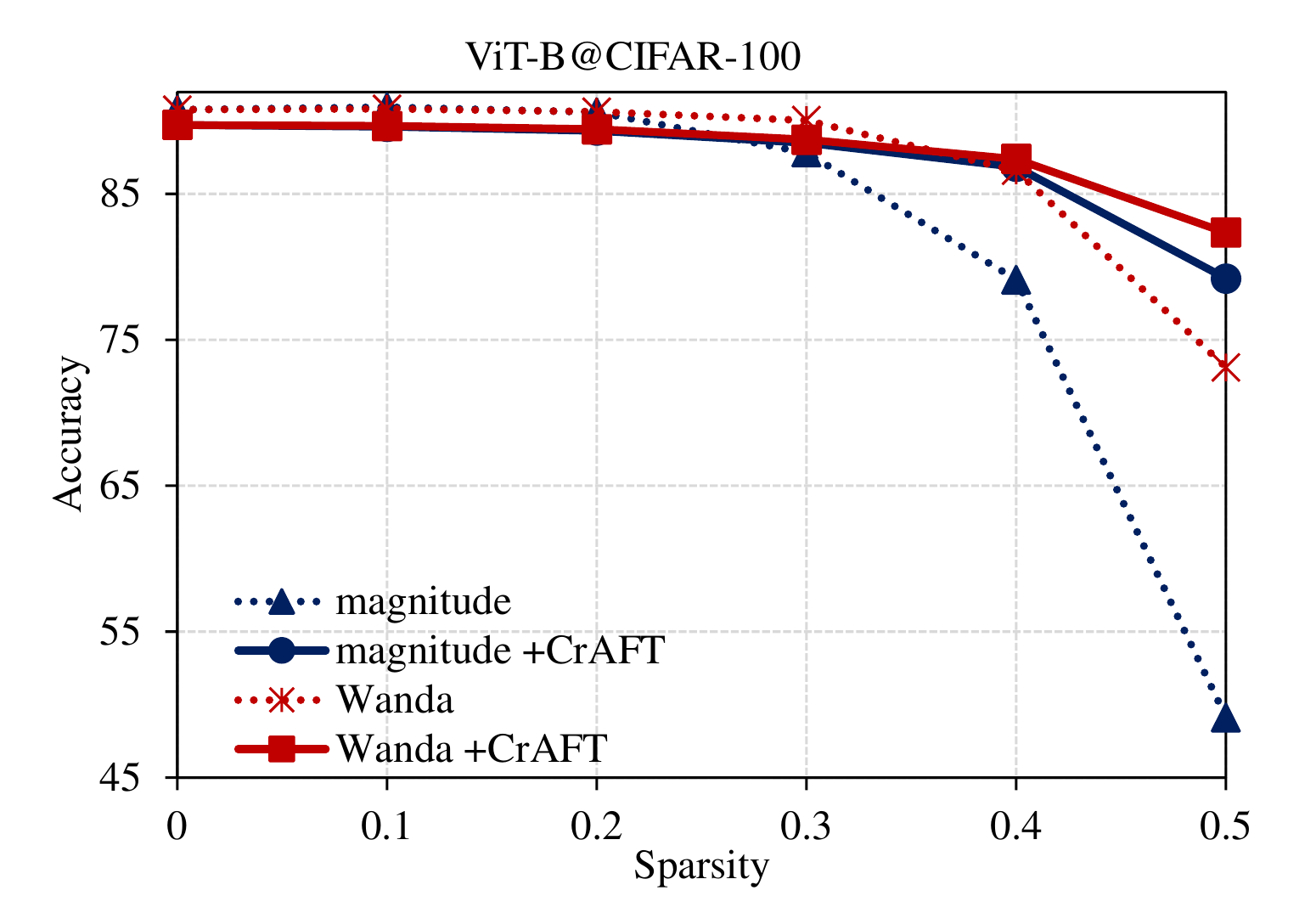}
      \caption{Unstructured Pruning}
      \label{fig:prune_a}
    \end{subfigure}%
    \begin{subfigure}{0.45\textwidth}
      \centering
      \includegraphics[width=\textwidth]{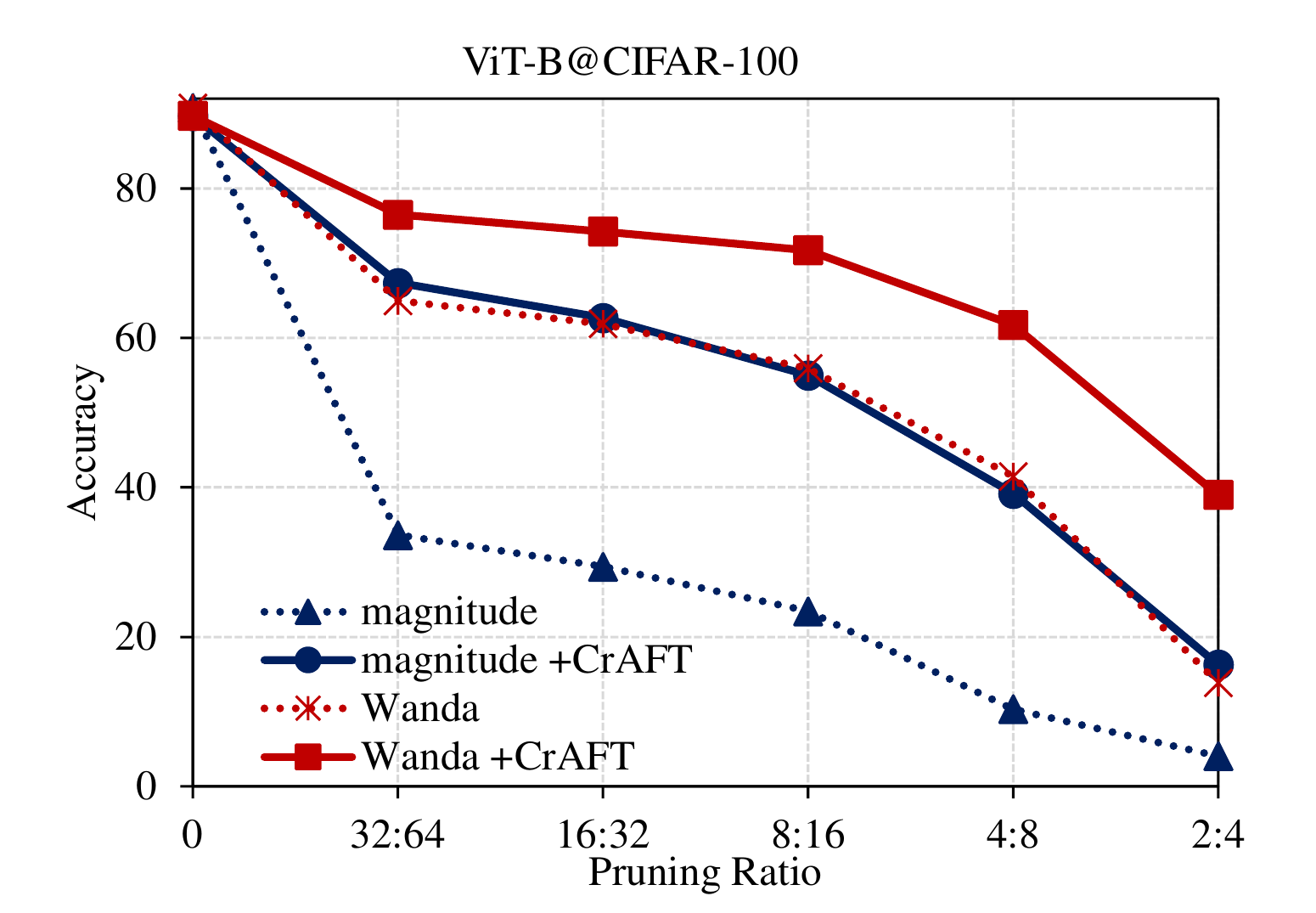}
      \caption{N:M Block pruning}
      \label{fig:prune_b}
    \end{subfigure}
\caption{\small \textbf{One-shot Weight Pruning.} Results of ViT-B@IMNET-21k fine-tuned on CIFAR100. As a simple baseline, we use magnitude-based pruning. More sophisticated approach, Wanda \cite{sun2023simple} employs activation-aware pruning to account for activation outliers. We report absolute top1 accuracy (\%) evaluated on both unstructured pruning and N:M block pruning (fixed at 50\% sparsity). CrAFT consistently provides accuracy lifts to both methods, achieving up to \textcolor{myemerald!80}{\textbf{12\%}} boost on 16:32.}
\label{fig:results:pruning}
\vspace{-8pt}
\end{figure*}
\myparagraph{Architecture Generalization Results.}
As in Table~\ref{tab:results:backbone}, CrAFT is able to generalize its efficacy across various foundation model architectures. When applying uniform symmetric quantization with BasePTQ, it consistently improves quantization results in both weight/activation and weight-only configurations. Its strongest gain is in the ViT-B backbone, where the W8A8 setting destroys the ADAM fine-tuned model to 54\% accuracy but CrAFT maintains at 81\% accuracy. This is relatedly consistent with the observation from Table~\ref{tab:motiv:sam}, where A8W8 also causes -72\% degradation. This phenomenon could be ascribed to the activation outliers problem as demonstrated in Figure~\ref{fig:motiv-quant}, where SAM indirectly smoothes the activation surface, which reduces the dynamic range and, therefore, quantization difficulty.

\subsubsection{One-Shot Pruning}

\paragraph{Baselines.}

As a strong baseline, we evaluate magnitude-based pruning \cite{han2015deep, frankle2018lottery}, which removes the lowest magnitude weights from the network. A more sophisticated method is Wanda \cite{sun2023simple}, which uses activation magnitudes to rescale the weight importance. The premise of this approach is that channels with large activations are more important; thus, the weight importance of those channels should be increased (therefore scaled up).

\myparagraph{Unstructured Pruning Results.}

As in Figure~\ref{fig:prune_a}, CrAFT achieves consistent performance improvements over both magnitude-based and activation-aware (Wanda) methods with non-trivial performance gaps. The differences are pronounced in high sparsity regions ($>$40\%) where CrAFT is able to lift around 9\% and an astounding 30\% for Wanda and magnitude-based, respectively.

\myparagraph{Block Pruning Results.}

Compared to unstructured pruning, block pruning is a more difficult task for two reasons: a) it usually requires 50\% sparsity per block (or group), which is a high degradation zone as seen in Figure~\ref{fig:prune_a}; b) it has a locality constraint, where least important weights are determined within each block (which may not be a globally optimal solution). For such reasons, the benefits of CrAFT are more pronounced as in Figure~\ref{fig:prune_b}. The difficulty of each N:M setting is in ascending order from left to right, as a bigger group size means less locality constraint. 

Surprisingly, ADAM baseline in the magnitude-based setting significantly drops in accuracy right from the easiest 32:64 setting, while CrAFT can maintain a 34\% accuracy point higher. Another interesting observation is the close alignment in the Pareto curve between Wanda and magnitude$+$CraFT. Note that these two curves are pruned from two \textit{separately fine-tuned networks} while the performance seems to be closely aligned. One could speculate that the sparse networks of these two settings are similar, where the potential benefit gained through CrAFT is equivalent to the importance of activation-aware weight scaling through Wanda.
\section{Analysis}
\label{sect:analysis}

In this section, we present analyses to examine the practicality of our method, as well as its sharpness-minimizing behavior during fine-tuning to better understand its underlying mechanics. All experiments are conducted with the default ViT-B@IMNET-21k model. 

\subsection{Training Overhead}

\begin{table}[!h]
    \setlength{\tabcolsep}{6pt}
    \small\centering
    \caption{\textbf{Analysis of the End to End (E2E) Fine-tuning time} for transfer learning experiments. The layout corresponds to Table~\ref{tab:results:transfer}, thus we encourage readers to cross-reference the runtime for each fine-tuning configuration. As expected, the sharpness-minimization objective in CrAFT adds around 2$\times$ overhead. However, the E2E time is still under \textit{a few minutes or hours on a single GPU}, which is a manageable load for fine-tuning vision foundation models. }
    
    \begin{tabular}{lccc}
        \toprule
    Downstream Task & Method & E2E Time (Min) & Overhead \\
    
    \midrule
    \multirow{2}{*}{Flowers-102} 
    & ADAM & 1.5 & 1.0$\times$\\
    &\cellcolor{gray!15} CrAFT 
    &\cellcolor{gray!15} 3.5 
    &\cellcolor{gray!15} 2.3$\times$\\

    \midrule
    \multirow{2}{*}{Stanford Cars} 
    & ADAM & 10.9 & 1.0$\times$\\
    &\cellcolor{gray!15} CrAFT 
    &\cellcolor{gray!15} 22  
    &\cellcolor{gray!15} 2.0$\times$\\

    \midrule
    \multirow{2}{*}{Oxford Pets} 
    & ADAM & 6.3 & 1.0$\times$\\
    &\cellcolor{gray!15} CrAFT &\cellcolor{gray!15} 14 &\cellcolor{gray!15} 2.2$\times$\\

    \midrule
    \multirow{2}{*}{Aircraft} 
    & ADAM & 10.4 & 1.0$\times$\\
    &\cellcolor{gray!15} CrAFT &\cellcolor{gray!15} 20  &\cellcolor{gray!15} 2.0$\times$\\

    \midrule
    \multirow{2}{*}{Food} 
    & ADAM & 85.2 & 1.0$\times$\\
    &\cellcolor{gray!15} CrAFT &\cellcolor{gray!15} 172 &\cellcolor{gray!15} 2.0$\times$\\

    \midrule
    \multirow{2}{*}{CIFAR} 
    & ADAM & 29  & 1.0$\times$\\
    &\cellcolor{gray!15} CrAFT &\cellcolor{gray!15} 57 &\cellcolor{gray!15} 1.9$\times$\\

        \bottomrule
    \end{tabular}
    \label{tab:analysis:time}
    \vspace{-8pt}
\end{table}


As aforementioned in Section~\ref{subsec:optim-difficult}, sharpness-minimization objectives, in our case, the vanilla SAM optimizer, can add considerable training overhead. In observation of such an optimization challenge, we seek to quantify the real wall-clock overhead from using CrAFT over the ADAM baseline. As seen in Table~\ref{tab:analysis:time}, CrAFT causes around 2$\times$ overhead in terms of the total number of minutes to complete the fine-tuning. Although this can mean considerable cost in large pretraining scenarios, we can get away with overheads in the order of a few minutes or hours on a single GPU, thanks to the quick fine-tuning schedule. 

\subsection{Sharpness Behaviors in Fine-Tuning}

\begin{table}[!h]
    \setlength{\tabcolsep}{6pt}
    \small\centering
    \caption{\textbf{Maximum curvature ($\lambda_\mathrm{max}$) of the loss surface} before and after fine-tuning for ViT-B@IMNET-21K. Before FT is equivalent to the converged base model after pre-training. Lower $\lambda_\mathrm{max}$ indicates flatter loss, which correlates with compressibility. CrAFT consistently achieves flatter losses compared to ADAM, which lays the foundation for superior quantization and pruning performances.}
    
    \begin{tabular}{lccc}
        \toprule
    Downstream Task & Before FT & ADAM & \cellcolor{gray!15} CrAFT \\
    
    \midrule
    Stanford Cars &
    125.63 & 103.97 & \cellcolor{gray!15} \textbf{84.22} \\

    \midrule
    Oxford Pets &
    284.22 & 135.6 & \cellcolor{gray!15} \textbf{61.05} \\

    \midrule
    Aircraft &
    172.72 & 142.93 & \cellcolor{gray!15} \textbf{109.35} \\
    
        \bottomrule
    \end{tabular}
    \label{tab:analysis:lambda}
    \vspace{-8pt}
\end{table}

Compressibility is correlated with the generalizability of the trained models. To this extent, it is important to quantify the measure of such qualities during the entire fine-tuning lifecycle. As in Table~\ref{tab:analysis:lambda}, we see that the pre-trained base models are relatively placed at a sharp minimum with respect to the downstream dataset. This means if compression is applied to the base model directly and evaluated on the target task in a zero-shot manner, it would face serious performance drops. 

Thankfully, once the fine-tuning has finished, the model can effectively find a flatter minimum that is also more robust to post-training compression. In comparing ADAM and our CrAFT methods, the latter method noticeably decreases the sharpness by up to 2$\times$ more in the Oxford Pets dataset. This is due to the sharpness minimization objective that helps the model to explicitly minimize the sharpness while also adapting to the target task.
\section{Related Works} \label{sec:prelim}

\subsection{Model Compression} \label{subsec:compression}
\subsubsection{Network Pruning}
\noindent \textbf{Weight Pruning.} Dating back to work in Optimal Brain Damage \cite{lecun1989optimal}, neural networks have been shown to contain redundant parameters that can be pruned without losing their performance. An iterative version of this was introduced \cite{han2015deep, frankle2018lottery} as an Iterative Magnitude Pruning (IMP) method, where the process of training and pruning repeats in cycles to obtain a sparse subnetwork. 

For better hardware efficiency, structured pruning \cite{he2017channel, mishra2021accelerating, lin2017runtime, heo2022sparse} seeks to trade pruning granularity for computational regularity. Our work benchmark unstructured weight pruning with one-shot methods since they do not require expensive retraining.


\subsubsection{Network Quantization}
\textbf{Post-training Quantization.}
The goal of quantization is to reduce the memory footprint and compute cost by reducing the bit widths of the network's numerical representations, including its weights, activations, and gradients. A naive baseline is a linear quantization, where the values are mapped linearly to quantization buckets. Post-training quantization (PTQ) involves approximating the activation distribution of a pre-trained model using some calibration data (e.g., a small subset of training data) to determine the clipping ranges and scaling factors \cite{DBLP:journals/corr/abs-2103-13630}. The model can then be statically quantized based on the activation statistics. More sophisticated approach use entropy-based \cite{park2017weighted} or piecewise linear functions \cite{fang2020post}.

Since quantization difficulties can differ across network components, another line of research deals with the granularity of quantization (per tensor, per channel, per token) or multi-bit quantization over different layers \cite{he2018amc,xiao2022smoothquant}. In our work, we benchmark INT8, INT6, and weight-only quantization.

\subsection{Vision Architectures} \label{subsec:backbone}
\textbf{Convolution-based.} CNNs have become a dominant way for computer vision research with the introduction of residual networks \cite{he2016deep}. More sophisticated networks have since been developed through efficient convolution primitives such as point-wise \cite{hua2018pointwise} and depth-wise \cite{chollet2017xception}. However, such parameter efficiency leads to difficulty in compression. Thus, we choose ResNet20 \cite{he2016deep} and MobileNetV1 \cite{howard2017mobilenets} networks as the model of choice to test convolution-based architectures across varying compression difficulties.

\textbf{Transformer-based.} ViTs \cite{dosovitskiy2020image} have emerged as a popular alternative to convolutional neural networks. In particular, the self-attention mechanism acts as a message-passing module between the tokens, which increases expressivity and lowers inductive bias. More sophisticated ViTs, such as Swin transformers \cite{liu2021swin}, use a hierarchical backbone to induce attention over local windows. 
Since Transformer models are most effective on larger datasets, we conduct the compressibility study on it but do not test architectural changes due to high training cost. 


\subsection{Sharpness Aware Minimization}\label{subsec:sam}
Researchers recently increasingly studied the curvature of the loss landscape and how it is linked to generalization in deep neural models. A notable study by Wu et al. \cite{DBLP:journals/corr/WuZE17} demonstrated that wide loss basins indicate simpler solutions. It has also been observed empirically that when directly targeting solutions in flat minima, generalization improves on various supervised learning tasks, including vision tasks.

Reference \cite{hochreiter1997long} introduced the idea that a flatter local minimum may improve the generalization performance of neural networks. Many subsequent works have explored the connection between loss geometry and generalization; for example, the Sharpness-Aware Minimizer (SAM) \cite{foret2020sharpness} explicitly minimizes the sharpness of the loss surface during training. However, due to the additional second-order computations, the training cost (and time) can go up by 2$\times$ \cite{duefficient}. Recently, more efficient alternatives \cite{duefficient,dusharpness,liu2022towards,pmlr-v139-kwon21b} have been proposed to reduce the training overhead. We choose the vanilla SAM optimizer as the flagship representative of the sharpness-minimizing methods to evaluate the relationship between the loss surface curvature and compressibility. 

\subsection{Compression-Aware Training}\label{subsec:pretrain+comp}
Compression-aware training is a severely under-explored area for model compression research due to its high computational complexity. We describe a few works that have attempted to build compressible models from scratch (pretraining) or from a pretrained base model (fine-tuning). 

\subsubsection{Pretraining}
\textbf{Pruning.}
Motivated by the fact that conventional pruning approaches require multiple retraining episodes with high computational cost, some works have sought to prune in a \emph{one-shot} manner (i.e., without any fine-tuning) by using sparsity-friendly pretraining methods. For example, CrAM \cite{peste2022cram} is a compression-aware minimizer that uses gradient ascent and approximated gradients to optimize against the perturbations that cause the most loss during training. Another method is SFW \cite{miao2021learning}, which uses the stochastic Frank-Wolfe algorithm to solve a constrained optimization problem by casting sparsity as a K-sparse polytope constraint.

\textbf{Quantization.}
SAQ \cite{liu2021sharpness} is a sharpness-aware quantization scheme that recognizes the existence of sharp discontinuities in the loss surface due to the discrete nature of low-precision training. In order to suppress unstable gradients, it proposes a SAM-style loss term that smoothes the loss surface to stabilize the training. In a similar vein, other works address the training instability with a closed-form normalization factor, e.g., \cite{lin2022device}. Both methods focus on the trainability challenges of quantization-aware training, not necessarily the compressibility of post-training quantized models. In our work, we overcome the trainability challenge by pretraining in full precision model with a vanilla SAM and applying quantization-aware post-training.

\subsubsection{Fine-tuning}
There has been even fewer works exploring fine-tuning for network compression. Closest to our work is in the natural language processing domain, where base models such as BERT have been explored to see the effect of sharpness minimization on compression \cite{na2022train}. 

However, our contributions make our work \textit{fundamentally different} in that: a) we include \textbf{empirical analysis} to motivate why sharpness minimization is favorable to pruning (Section~\ref{subsec:prune-difficult}) and quantization (Section~\ref{subsec:quant-difficult}), and b) we for the first time explore vision foundation models, which have \textbf{vision-specific compression techniques} such as PTQ4ViT which show strong results when combined with our CrAFT framework.
\section{Conclusion}
This paper introduces CrAFT to close the performance gap of conventional vision foundation models, which are fine-tuned without compressibility in mind. Through an in-depth analysis of compression difficulties and their relationship to loss sharpness, we propose to minimize the sharpness during fine-tuning. We demonstrate the strong empirical advantages of adopting CrAFT with existing compression techniques such as one-shot pruning and post-training quantization. We hope that CrAFT can inspire future research on studying compression through the lens of generalizability (or robustness to adversarial perturbations).

\paragraph{Limitations and Future Works.}
We employ a vanilla implementation of the SAM optimizer to examine the baseline performance of such sharpness minimization approaches for model compression. Since the training overhead can quickly become intractable for larger foundation models, we believe more parameter efficient (e.g. LoRA \cite{hu2021lora}) versions of CrAFT could be useful for scaling up foundation models.

{\small
\bibliographystyle{ieee_fullname}
\bibliography{egbib}
}


\end{document}